\newif\ifLNAI
\LNAItrue
\newif\ifCompact
\ifLNAI
\Compactfalse
\else
\Compacttrue
\fi
\newif\ifllncs
\ifLNAI
\llncstrue
\else
\llncstrue
\fi
\ifCompact
  \ifllncs
    \documentclass[twocolumn,10pt,runningheads, envcountsame]{llncs}
  \else
    \documentclass[twolumn,10pt]{article}
  \fi
\else
  \documentclass[runningheads, envcountsame, a4paper]{llncs}
\fi
\usepackage{microtype}
\usepackage{amsfonts}

\usepackage{graphicx} 
\usepackage{subfigure}
\usepackage[utf8]{inputenc}
\usepackage{natbib}

\title{Deep Learning of Representations:\\ Looking Forward}
\titlerunning{Deep Learning of Representations: Looking Forward}
\toctitle{Deep Learning of Representations: Looking Forward}
\author{Yoshua Bengio}
\authorrunning{Y. Bengio}
\tocauthor{Yoshua~Bengio}
\institute{Department of Computer Science and Operations Research\\
Université de Montréal, Canada}

\newcommand{\E}{\mathbb{E}}

\date{}

\begin{document}
\maketitle
\begin{abstract}
Deep learning research aims at discovering learning algorithms that discover multiple levels of distributed representations, with higher levels representing more abstract concepts. Although the study of deep learning has already led to impressive theoretical results, learning algorithms and breakthrough experiments, several challenges lie ahead. This paper proposes to examine some of these challenges, centering on the questions of scaling deep learning algorithms to much larger models and datasets, reducing optimization difficulties due to ill-conditioning or local minima, designing more efficient and powerful inference and sampling procedures, and learning to disentangle the factors of variation underlying the observed data.  It also proposes a few forward-looking research directions aimed at overcoming these challenges.

\end{abstract}

\section{Background on Deep Learning}

Deep learning is an emerging approach within the machine learning research
community. Deep learning algorithms have been proposed in recent years to
move machine learning systems towards the discovery of multiple levels of
representation.  They have had important empirical successes in a number of
traditional AI applications such as computer vision and natural language
processing.
See~\citep{Bengio-2009-book,Bengio-Courville-Vincent-TPAMI2013} for reviews
and~\citet{Bengio-tricks-chapter-2013} and the other chapters of the
book by~\citet{Montavon2012} for practical guidelines.  Deep learning is
attracting much attention both from the academic and industrial
communities.  Companies like Google, Microsoft, Apple, IBM and Baidu are
investing in deep learning, with the first widely distributed products
being used by consumers aimed at speech recognition.  Deep learning is also
used for object recognition (Google Goggles), image and music information
retrieval (Google Image Search, Google Music), as well as computational
advertising~\citep{Corrado-2012}.
A deep learning
building block (the {\em restricted Boltzmann machine}, or RBM) was used as a
crucial part of the winning entry of a million-dollar machine learning
competition (the Netflix
competition)~\citep{SalakhutdinovR2007b-short,BigChaos-Netflix2009}. 
The New York Times covered the subject twice in 2012, with
front-page articles.\footnote{\raggedright{\parindent=1.5em\url{
http://www.nytimes.com/2012/11/24/science/scientists-see-advances-
in-deep-learning-a-part-of-artificial-intelligence.html}}}
Another series of articles (including a third New York Times article)
covered a more recent event showing off the application of deep learning in
a major Kaggle competition for drug discovery (for example see ``Deep
Learning - The Biggest Data Science Breakthrough of the Decade''\footnote{\url
{http://oreillynet.com/pub/e/2538}}.  Much
more recently, Google bought out (``acqui-hired'') a company (DNNresearch)
created by University of Toronto professor Geoffrey Hinton (the founder and
leading researcher of deep learning) and two of his PhD students, Ilya
Sutskever and Alex Krizhevsky, with the press writing titles such as
``Google Hires Brains that Helped Supercharge Machine Learning'' (Robert
McMillan for Wired, March 13th, 2013).

The performance of many machine learning methods is heavily dependent on
the choice of data representation (or features) on which they are applied.
For that reason, much of the actual effort in deploying machine learning
algorithms goes into the design of preprocessing pipelines that result in a
hand-crafted representation of the data that can support effective machine
learning.  Such feature engineering is important but labor-intensive and
highlights the weakness of many traditional learning algorithms: their
inability to extract and organize the discriminative information from the
data.  Feature engineering is a way to take advantage of human ingenuity
and prior knowledge to compensate for that weakness. In order to expand the
scope and ease of applicability of machine learning, it would be highly
desirable to make learning algorithms less dependent on feature
engineering, so that novel applications could be constructed faster, and
more importantly for the author, to make progress towards artificial
intelligence (AI).

A representation learning algorithm discovers explanatory factors or
features. A deep learning algorithm is a particular kind of representation
learning procedure that discovers {\em multiple levels of representation,
with higher-level features representing more abstract aspects of the data}.
This area of research was kick-started in 2006 by a few research groups,
starting with Geoff Hinton's group, who initially focused on stacking
unsupervised representation learning algorithms to obtain deeper
representations~\citep{Hinton06,Bengio-nips-2006-small,ranzato-07-small,HonglakL2008-small}.
Since then, this area has seen rapid growth, with an increasing number of
workshops (now one every year at the NIPS and ICML conferences, the two
major conferences in machine learning) and even a new specialized
conference just created in 2013 (ICLR -- the International Conference on
Learning Representations).

Transfer learning is the ability of a learning algorithm to exploit
commonalities between different learning tasks in order to share
statistical strength, and {\em transfer knowledge} across tasks. Among the
achievements of unsupervised representation learning algorithms are the
impressive successes they obtained at the two transfer learning challenges
held in 2011. First, the Transfer Learning Challenge, presented at an ICML
2011 workshop of the same name, was won using unsupervised layer-wise
pre-training~\citep{UTLC+DL+tutorial-2011-small,UTLC+LISA-2011-small}. A
second Transfer Learning Challenge was held the same year and won
by~\citet{Goodfellow+all-NIPS2011} using unsupervised representation
learning.  Results were presented at NIPS 2011's Challenges in Learning
Hierarchical Models Workshop.

\section{Quick Overview of Deep Learning Algorithms}

The central concept behind all deep learning methodology is the automated discovery
of abstraction, with the belief that more abstract representations of data
such as images, video and audio signals tend to be more useful: they
represent the semantic content of the data, divorced from the low-level
features of the raw data (e.g., pixels, voxels, or waveforms). Deep
architectures lead to abstract representations because more abstract
concepts can often be constructed in terms of less abstract ones.

Deep learning algorithms are special cases of representation learning with
the property that they learn multiple levels of representation. Deep
learning algorithms often employ shallow (single-layer) representation
learning algorithms as subroutines. Before covering the unsupervised
representation learning algorithms, we quickly review the basic principles
behind supervised representation learning algorithms such as the good old
multi-layer neural networks. Supervised and unsupervised objectives can of
course be combined (simply added, with a hyper-parameter as coefficient),
like in~\citet{Larochelle+Bengio-2008-small}'s discriminative RBM.

\subsection{Deep Supervised Nets, Convolutional Nets, Dropout}
\label{sec:dnn}

Before 2006, it was believed that training deep supervised neural
networks~\citep{Rumelhart86b-small} was too difficult (and indeed did not
work). The first breakthrough in training them happened in Geoff Hinton's
lab with unsupervised pre-training by RBMs~\citep{Hinton06}, as discussed
in the next subsection. However, more recently, it was discovered that one
could train deep supervised nets by proper initialization, just large
enough for gradients to flow well and activations to convey useful
information~\citep{GlorotAISTATS2010-small,Sutskever-thesis2012-small}.\footnote{and
potentially with the use of momentum~\citep{Sutskever-thesis2012-small}} Another
interesting ingredient in the success of training the deep supervised
networks of~\citet{GlorotAISTATS2010-small} (and later
of~\citet{Krizhevsky-2012-small}) is the presence of rectifying
non-linearities (such as $\max(0,x)$) instead of sigmoidal non-linearities
(such as $1/(1+\exp(-x))$ or $\tanh(x)$).
See~\citet{Jarrett-ICCV2009-small,Hinton2010} for earlier work on
rectifier-like non-linearities. We return to this topic in
Section~\ref{sec:optimization}.  These good results with purely
supervised training of deep nets seem to be especially clear when large
quantities of labeled data are available, and it was demonstrated with
great success for speech
recognition~\citep{Seide2011,Hinton-et-al-2012,LiDeng-et-al-ICASSP-2013-small}
and object recognition~\citep{Krizhevsky-2012-small} with breakthroughs
reducing the previous state-of-the-art error rates by 30\% to 50\% on
difficult to beat benchmarks.

One of the key ingredients for success in the applications of deep learning
to speech, images, and natural language
processing~\citep{Bengio-scholarpedia-2007-small,collobert:2011b} is the
use of {\em convolutional} architectures~\citep{LeCun98-small}, which
alternate {\em convolutional layers} and {\em pooling layers}. Units on
hidden layers of a convolutional network are associated with a spatial or
temporal position and only depend on (or generate) the values in a
particular window of the raw input. Furthermore, units on convolutional
layers share parameters with other units of the same ``type'' located at
different positions, while at each location one finds all the different
types of units. Units on pooling layers aggregate the outputs of units at a
lower layer, either aggregating over different nearby spatial positions (to
achieve a form of local spatial invariance) or over different unit
types. For example, a {\em max-pooling} unit outputs the maximum over some
lower level units, which can therefore be seen to compete towards sending
their signal forward.

Another key ingredient in the success of many recent breakthrough results
in the area of object recognition is the idea of {\em
dropouts}~\citep{Hinton-et-al-arxiv2012,Krizhevsky-2012-small,Goodfellow+al-ICML2013-small}.
Interestingly, it consists in {\em injecting noise} (randomly dropping out
units with probability $1/2$ from the neural network during training, and
correspondingly multiplying by $1/2$ the weights magnitude at test time)
that prevents a too strong co-adaptation of hidden units: hidden units must
compute a feature that will be useful even when half of the other hidden
units are stochastically turned off (masked). This acts like a powerful
regularizer that is similar to bagging aggregation but over an
exponentially large number of models (corresponding to different masking
patterns, i.e., subsets of the overall network) that share parameters.

\subsection{Unsupervised or Supervised Layer-wise Pre-Training}

One of the key results of recent years of research in deep learning is that
deep compositions of non-linearities -- such as found in deep feedforward
networks or in recurrent networks applied over long sequences -- can be
very sensitive to initialization (some initializations can lead much better
or much worse results after training).  The first type of approaches that
were found useful to reduce that sensitivity is based on {\em greedy
layer-wise pre-training}~\citep{Hinton06,Bengio-nips-2006-small}.  The idea
is to train one layer at a time, starting from lower layers (on top of the
input), so that there is a clear training objective for the currently added
layer (which typically avoids the need for back-propagating error gradients
through many layers of non-linearities).  With unsupervised pre-training,
each layer is trained to model the distribution of values produced as
output of the previous layer. As a side-effect of this training, a new
representation is produced, which can be used as input for deeper layers.
With the less common {\em supervised}
pre-training~\citep{Bengio-nips-2006-small,Yu+al-2010,Seide-et-al-ASRU2011},
each additional layer is trained with a supervised objective (as part of a
one hidden layer network).  Again, we obtain a new representation (e.g.,
the hidden or output layer of the newly trained supervised model) that can
be re-used as input for deeper layers.  The effect of unsupervised
pre-training is apparently most drastic in the context of training deep
{\em auto-encoders}~\citep{Hinton-Science2006}, unsupervised learners that
learn to reconstruct their input: unsupervised pre-training allows to find
much lower training and test reconstruction error.

\subsection{Directed and Undirected Graphical Models with Anonymous Latent Variables}

Anonymous latent variables are latent variables that do not have a
predefined semantics in terms of predefined human-interpretable concepts.
Instead they are meant as a means for the computer to discover underlying
explanatory factors present in the data. We believe that although
non-anonymous latent variables can be very useful when there is sufficient
prior knowledge to define them, anonymous latent variables are very useful
to let the machine discover complex probabilistic structure: they lend
flexibility to the model, allowing an otherwise parametric model to
non-parametrically adapt to the amount of data when more anonymous
variables are introduced in the model.

{\em Principal components analysis} (PCA), {\em independent components analysis} (ICA),
and sparse coding all correspond to a directed graphical model in which the observed vector $x$
is generated by first independently sampling some underlying factors (put in vector $h$)
and then obtaining $x$ by $W h$ plus some noise. They only differ in the type of
prior put on $h$, and the corresponding {\em inference} procedures to recover
$h$ (its posterior $P(h\mid x)$ or expected value $\E[h\mid x]$) when $x$ is observed.
Sparse coding tends to yield many zeros in the estimated vector $h$ that could
have generated the observed $x$.
See section 3 of~\citet{Bengio-Courville-Vincent-TPAMI2013} for a review of
representation learning procedures based on directed or undirected
graphical models.\footnote{Directed and undirected: just
two different views on the semantics of probabilistic models, not mutually
exclusive, but views that are more convenient for some models than others.}
Section~\ref{sec:sparse-coding} describes sparse coding in more detail.

An important thing to keep in mind is that directed graphical models tend
to enjoy the property that in computing the posterior, the different factors
{\em compete with each other}, through the celebrated {\em explaining away effect}.
Unfortunately, except in very special cases (e.g., when the columns of $W$
are orthogonal, which eliminates explaining away and its need), 
this results in computationally expensive inference.
Although {\em maximum a posteriori} (MAP) inference\footnote{finding $h$ that
approximately maximizes $P(h\mid x)$}
 remains polynomial-time in the case of sparse coding,
this is still very expensive, and unnecessary in other types of models (such as the stacked
auto-encoders discussed below). In fact, exact inference becomes intractable
for deeper models, as discussed in section~\ref{sec:inference}.

Although RBMs enjoy tractable inference, this is obtained at the cost
of a lack of explaining away between the hidden units,
which could potentially limit the representational power of $\E[h\mid x]$
as a good representation for the factors that could have generated $x$.
However, RBMs are often used as building blocks for training deeper
graphical models such as the {\em deep belief network}~(DBN)~\citep{Hinton06} and
sthe {\em deep Boltzmann machine}~(DBM)~\citep{SalHinton09small}, which can
compensate for the lack of explaining away in the RBM hidden units
via a rich prior (provided by the upper layers) which can introduce
potentially complex interactions and competition between the hidden units.
Note that there is explaining away (and intractable exact inference)
iin DBNs and something analogous in DBMs.

\subsection{Regularized Auto-Encoders}

Auto-encoders include in their training criterion a form of reconstruction
oerror, such as $||r(x)-x||^2$, where $r(\cdot)$ is the learned
reconstruction function, often decomposed as $r(x)=g(f(x))$ where
$f(\cdot)$ is an encoding function and $g(\cdot)$ a decoding function.  The
idea is that auto-encoders should have low reconstruction error at the
training examples, but high reconstruction error in most other
configurations of the input. In the case of auto-encoders, good
generalization means that test examples (sampled from the same distribution
as training examples) also get low reconstruction error.  Auto-encoders
have to be regularized to prevent them from simply learning the identity
function $r(x)=x$, which would be useless.  {\em Regularized auto-encoders}
include the old bottleneck auto-encoders (like in PCA) with less hidden
units than input, as well as the denoising
auto-encoders~\citep{VincentPLarochelleH2008-small} and contractive
auto-encoders~\citep{Rifai+al-2011-small}. The denoising auto-encoder takes
a noisy version $N(x)$ of original input $x$ and tries to
reconstruct $x$, e.g., it minimizes $||r(N(x))-x||^2$.  The contractive
auto-encoder has a regularization penalty in addition to the reconstruction
error, trying to make hidden units $f(x)$ as constant as possible with
respect to $x$ (minimizing the contractive penalty $||\frac{\partial
f(x)}{\partial x}||^2_F$).  A Taylor expansion of the denoising error shows
that it is also approximately equivalent to minimizing reconstruction error plus a
contractive penalty on $r(\cdot)$ \citep{Alain+Bengio-ICLR2013}.  As
explained in~\citet{Bengio-Courville-Vincent-TPAMI2013}, the tug-of-war
between minimization of reconstruction error and the regularizer means that
the intermediate representation must mostly capture the variations
necessary to distinguish training examples, i.e., the directions of
variations on the {\em manifold} (a lower dimensional region) near which
the data generating distribution concentrates. {\em Score
matching}~\citep{Hyvarinen-2005-small} is an inductive principle that can
be an interesting alternative to maximum likelihood, and several
nconnections have been drawn between reconstruction error in auto-encoders
and score matching~\citep{Swersky-ICML2011}.  It has also been shown that
denoising auto-encoders and some forms of contractive auto-encoders
estimate the score\footnote{derivative of the log-density with respect to
the data; this is different from the usual definition of score in
statistics, where the derivative is with respect to the parameters} of the
underlying data generating
distribution~\citep{Vincent-NC-2011,Alain+Bengio-ICLR2013}.  This can be used
to endow regularized auto-encoders with a probabilistic interpretation and
to sample from the implicitly learned density
models~\citep{Rifai-icml2012-small,Bengio-arxiv-moments-2012,Alain+Bengio-ICLR2013}
through some variant of Langevin or Metropolis-Hastings {\em Monte-Carlo
Markov chains} (MCMC). More recently, the results from \citet{Alain+Bengio-ICLR2013}
have been generalized: whereas the score estimation result was only
valid for asymptotically small Gaussian corruption noise, squared
reconstruction error, and continuous inputs, the result from \citet{Bengio-et-al-arxiv-2013}
is applicable for any type of input, any form of reconstruction loss
(so long as it is a negative log-likelihood), any form of corruption
(so long as it prevents learning the identity mapping) and does not
depend on the level of corruption noise going to zero.

Even though there is a probabilistic interpretation to regularized
auto-encoders, this interpretation does not involve the definition of
intermediate anonymous latent variables. Instead, they are based on the
construction of a direct parametrization of an encoding function which
immediately maps an input $x$ to its representation $f(x)$, and they are
motivated by geometrical considerations in the spirit of manifold learning
algorithms~\citep{Bengio-Courville-Vincent-TPAMI2013}.  Consequently, there
is no issue of tractability of inference, even with deep auto-encoders
obtained by stacking single-layer ones. This is true even in the
recently proposed multi-layer versions of the denoising 
auto-encoders~\citep{Bengio+Laufer-arxiv-2013}
in which noise is injected not just in input, but in hidden units
(like in the Gibbs chain of a deep Boltzmann machine).

It was previously
believed~\citep{ranzato-08}, including by the author himself, that
reconstruction error should only be small where the estimated density has a
peak, e.g., near the data. However, recent theoretical and empirical
results~\citep{Alain+Bengio-ICLR2013} show that the
reconstruction error will be small where the estimated density has a peak
(a mode) but also where it has a trough (a minimum).  This is because the
reconstruction error vector (reconstruction minus input) estimates the
score $\frac{\partial \log p(x)}{\partial x}$, i.e., the reconstruction
error is small where $||\frac{\partial \log p(x)}{\partial x}||$ is
small. This can happen at a local maximum but also at a local minimum (or
saddle point) of the estimated density. This argues against using
reconstruction error itself as an {\em energy}
function,\footnote{To define energy, we write probability as the
normalized exponential of minus the
energy.} which should only be low near high probability points.

\subsection{Sparse Coding and PSD}
\label{sec:sparse-coding}

Sparse coding~\citep{Olshausen+Field-1996} is a particular kind of directed
graphical model with a linear relationship between visible and latent
variables (like in PCA), but in which the latent variables have a prior
(e.g., Laplace density) that encourages sparsity (many zeros) in the MAP
posterior. Sparse coding is not actually very good as a generative model,
but has been very successful for unsupervised feature
learning~\citep{RainaR2007-small,Coates2011b,Yu+Lin+Lafferty-2011-short,Grosse-2007-small,Jenatton-2009,Bach2011-small}.
See~\citet{Bengio-Courville-Vincent-TPAMI2013} for a brief overview in the
context of deep learning, along with connections to other unsupervised
representation learning algorithms. Like other directed graphical models,
it requires somewhat expensive inference, but the good news is that for
sparse coding, MAP
inference is a convex optimization problem for which several fast
approximations have been
proposed~\citep{Mairal-et-al-ICML2009,Gregor+LeCun-ICML2010}.  It is
interesting to note the results obtained by~\citet{Coates2011b} which
suggest that sparse coding is a {\em better encoder but not a better
learning algorithm} than RBMs and sparse auto-encoders (none of which has
explaining away).  Note also that sparse coding can be generalized into the
spike-and-slab sparse coding algorithm~\citep{Goodfellow-icml2012}, in
which MAP inference is replaced by variational inference, and that was used
to win the NIPS 2011 transfer learning
challenge~\citep{Goodfellow+all-NIPS2011}.

Another interesting variant on sparse coding is the {\em predictive sparse
coding} (PSD) algorithm~\citep{koray-psd-08-small} and its variants, which
combine properties of sparse coding and of auto-encoders. Sparse coding can
be seen as having only a parametric ``generative'' decoder (which maps
latent variable values to visible variable values) and a non-parametric
encoder (find the latent variables value that minimizes reconstruction
error and minus the log-prior on the latent variable). PSD adds a
parametric encoder (just an affine transformation followed by a
non-linearity) and learns it jointly with the generative model, such that
the output of the parametric encoder is close to the latent variable values
that reconstructs well the input.

\section{Scaling Computations}

From a computation point of view,
how do we scale the recent successes of deep learning to much larger
models and huge datasets, such that the models are actually richer and capture
a very large amount of information?

\subsection{Scaling Computations: The Challenge}

The beginnings of deep learning in 2006 have focused on the MNIST digit
image classification problem~\citep{Hinton06,Bengio-nips-2006-small},
breaking the supremacy of SVMs (1.4\% error) on this dataset.\footnote{for
the knowledge-free version of the task, where no image-specific prior is
used, such as image deformations or convolutions, where the current
state-of-the-art is around 0.8\% and involves deep
learning~\citep{Dauphin-et-al-NIPS2011-small,Hinton-et-al-arxiv2012}.}
The latest records are still held by deep networks:
~\citet{Ciresan-2012}  currently claim the title of state-of-the-art for the unconstrained version of the task
(e.g., using a convolutional architecture and stochastically
deformed data), with 0.27\% error.

In the last few years, deep learning has moved from digits to object
recognition in natural images, and the latest breakthrough has been
achieved on the ImageNet dataset.\footnote{The 1000-class ImageNet
benchmark, whose results are detailed here:\\ {\tt\scriptsize
http://www.image-net.org/challenges/LSVRC/2012/
results.html}} bringing down
the state-of-the-art error rate (out of 5 guesses) from 26.1\% to
15.3\%~\citep{Krizhevsky-2012-small}

To achieve the above scaling from 28$\times$28 grey-level MNIST images to
256$\times$256 RGB images, researchers have taken advantage of
convolutional architectures (meaning that hidden units do not need to be
connected to all units at the previous layer but only to those in the same
spatial area, and that pooling units reduce the spatial resolution as we
move from lower to higher layers). They have also taken advantage of GPU
technology to speed-up computation by one or two orders of
magnitude~\citep{RainaICML09,bergstra+al:2010-scipy-short,bergstra+all-Theano-NIPS2011,Krizhevsky-2012-small}.

We can expect computational power to continue to increase, mostly through
increased parallelism such as seen in GPUs, multicore machines, and
clusters. In addition, computer memory has become much more affordable,
allowing (at least on CPUs) to handle potentially huge models (in terms of
capacity).

However, whereas the task of recognizing handwritten digits is solved
to the point of achieving roughly human-level performance, this is far
from true for tasks such as general object recognition, scene understanding,
speech recognition, or natural language understanding.  What is needed
to nail those tasks and scale to even more ambitious ones?

As we approach AI-scale tasks, it should become clear that our trained
models will need to be much larger in terms of number of parameters.  This
is suggested by two observations. First, AI means understanding the world
around us at roughly the same level of competence as humans. Extrapolating
from the current state of machine learning, the amount of knowledge this
represents is bound to be large, many times more than what current models
can capture. Second, more and more empirical results with deep learning
suggest that larger models systematically work
better~\citep{Coates2011-shorter,Hinton-et-al-arxiv2012,Krizhevsky-2012-small,Goodfellow+al-ICML2013-small},
provided appropriate regularization is used, such as the dropouts technique
described above.

Part of the challenge is that the current capabilities of a single computer
are not sufficient to achieve these goals, even if we assume that training complexity
would scale linearly with the complexity of the task. This has for example
motivated the work of the Google Brain
team~\citep{QuocLe-ICML2012-small,Dean-et-al-NIPS2012} to parallelize
training of deep nets over a very large number of nodes. As we will see in
Section~\ref{sec:optimization}, we hypothesize that as the size of the
models increases, our current ways of training deep networks become less
and less efficient, so that the computation required to train larger models
(to capture correspondingly more information) is likely to scale much worse
than linearly~\citep{Dauphin+Bengio-arxiv-2013}.

Another part of the challenge is that the increase in computational power
has been mostly coming (and will continue to come) from parallel computing.
Unfortunately, when considering very large datasets, our most efficient
training algorithms for deep learning (such as variations on {\em
stochastic gradient descent} or SGD) are inherently sequential (each update
of the parameters requires having completed the previous update, so they
cannot be trivially parallelized). Furthermore, for some tasks, the amount
of available data available is becoming so large that it does not fit on a
disk or even on a file server, so that it is not clear how a single CPU
core could even scan all that data (which seems necessary in order to learn
from it and exploit all of it, if training is inherently sequential).

\subsection{Scaling Computations: Solution Paths}

\subsubsection{Parallel Updates: Asynchronous SGD.}
One idea that we explored in~\citet{Bengio-nnlm2003-small} is that of {\em
asynchronous SGD}: train multiple versions of the model in parallel, each
running on a different node and seeing different subsets of the data (on
different disks), but with an asynchronous lock-free sharing mechanism which keeps
the different versions of the model not too far from each other.  If the
sharing were synchronous, it would be too inefficient because most nodes
would spend their time waiting for the sharing to be completed and would be
waiting for the slowest of the nodes. This idea has been analyzed
theoretically~\citep{Recht-et-al-NIPS2011} and successfully engineered on a
grand scale recently at
Google~\citep{QuocLe-ICML2012-small,Dean-et-al-NIPS2012}. However, current
large-scale implementations (with thousands of nodes) are still very
inefficient (in terms of use of the parallel resources), mostly because of
the communication bottleneck requiring to regularly exchange parameter
values between nodes. The above papers also take advantage of a way to
train deep networks which has been very successful for GPU implementations,
namely the use of rather large minibatches (blocks of examples after which
an update is performed), making some parallelization (across the examples
in the minibatch) easier. One option, explored by~\citet{Coates-et-al-NIPS2012}
is to use as building blocks for learning features algorithms such as  k-means
that can be run efficiently over large minibatches (or the whole data)
and thus parallelized easily on a cluster (they learned 150,000 features
on a cluster with only 30 machines).

Another interesting consideration is the optimization of trade-off between
communication cost and computation cost in distributed optimization
algorithms, e.g., as discussed in~\citet{Tsianos-et-al-NIPS2012-small}.

\subsubsection{Sparse Updates.}

One idea that we propose here is to change the learning algorithms so as to
obtain {\em sparse updates}, i.e., for any particular minibatch there is
only a small fraction of parameters that are updated.  If the amount of sparsity
in the update is large, this would mean that a much smaller fraction of the
parameters need to be exchanged between nodes when performing an
asynchronous SGD\footnote{although the gain would be reduced considerably
in a minibatch mode, roughly by the size of the minibatch}.  Sparse updates
could be obtained simply if the gradient is very sparse. This gradient
sparsity can arise with approaches that select paths in the neural
network. We already know methods which produce slightly sparse updates,
such as dropouts~\citep{Hinton-et-al-arxiv2012},\footnote{where half of the
hidden units are turned off, although clearly, this is not enough sparsity
for reaching our objective; unfortunately, we observed that randomly and
independently dropping a lot more than half of the units yielded
substantially worse results}
maxout~\citep{Goodfellow+al-ICML2013-small}\footnote{where in addition to
dropouts, only one out of $k$ filters wins the competition in max-pooling
units, and only one half of those survives the dropouts masking, making the
sparsity factor 2$k$} and other hard-pooling mechanisms, such as the
recently proposed and very successful stochastic
pooling~\citep{Zeiler+et+al-ICLR2013}. These methods do not provide enough
sparsity, but this could be achieved in two ways. First of all, we could
choose to only pay attention to the largest elements of the gradient
vector. Second, we could change the architecture along the lines proposed
next.

\subsubsection{Conditional Computation.}
A central idea (that applies whether one parallelizes or not) that we put
forward is that of {\em conditional computation}: instead of dropping out
paths independently and at random, drop them in a learned and optimized
way.  Decision trees remain some of the most appealing machine learning
algorithms because prediction time can be on the order of the logarithm of
the number of parameters. Instead, in most other machine learning
predictors, scaling is linear (i.e., much worse).  This is because decision
trees exploit conditional computation: for a given example, as additional
computations are performed, one can discard a gradually larger set of
parameters (and avoid performing the associated computation). In deep
learning, this could be achieved by combining {\em truly sparse
activations} (values not near zero like in sparse auto-encoders, but actual
zeros) and {\em multiplicative connections} whereby some hidden units {\em
gate} other hidden units (when the gater output is zero it turns off the
output of the gated unit). When a group A of hidden units has a sparse
activation pattern (with many actual zeros) and it multiplicatively gates
other hidden units B, then only a small fraction of the hidden units in B
may need to be actually computed, because we know that these values will
not be used.  Such gating is similar to what happens when a decision node
of a decision tree selects a subtree and turns off another subtree. More
savings can thus be achieved if units in B themselves gate other units,
etc. The crucial difference with decision trees (and e.g., the hard mixture
of experts we introduced a decade ago~\citep{Collobert+Bengio+Bengio-2003})
is that the gating units should {\em not be mutually exclusive} and should
instead form a {\em distributed pattern}. Indeed, we want to keep the
advantages of distributed representations and avoid the limited local
generalization suffered by decision
trees~\citep{Bengio-decision-trees10}. With a high level of conditional
computation, some parameters are used often (and are well tuned) whereas
other parameters are used very rarely, requiring more data to estimate. A
trade-off and appropriate regularization
therefore needs to be established which will depend on the amount of training
signals going into each parameter. Interestingly, {\em conditional computation also helps to
achieve sparse gradients}, and the fast convergence of hard mixtures of
experts~\citep{Collobert+Bengio+Bengio-2003} provides positive evidence
that a side benefit of conditional computation will be easier and faster
optimization.

Another existing example of conditional computation and sparse gradients
is with the first layer of neural language models, deep learning models
for text data~\citep{Bengio-nnlm2003-small,Bengio-scholarpedia-2007-small}.
In that case, there is one parameter vector per word in the vocabulary,
but each sentence only ``touches'' the parameters associated with the
words in the sentence. It works because the input can be seen as extremely
sparse. The question is how to perform conditional computation
in the rest of the model.

One issue with the other example we mentioned, hard mixtures of
experts~\citep{Collobert+Bengio+Bengio-2003}, is that its training
mechanism only make sense when the gater operates at the output layer. In
that case, it is easy to get a strong and clean training signal for the
gater output: one can just evaluate what the error would have been if a
different expert had been chosen, and train the gater to produce a higher
output for the expert that would have produced the smallest error (or to
reduce computation and only interrogate two experts, require that the gater
correctly ranks their probability of being the best one). The challenge is
how to produce training signals for gating units that operate in the middle
of the model. One cannot just enumerate all the gating configurations,
because in a distributed setting with many gating units, there will be an
exponential number of configurations.  Interestingly, this suggests {\em
  introducing randomness} in the gating process itself, e.g.,
stochastically choosing one or two choices out of the many that a group of
gating units could take. This is interesting because this is the
second motivation (after the success of dropouts as a regularizer)
for re-introducing randomness in the middle of deep networks. This randomness
would allow configurations that would otherwise not be selected (if
only a kind of ``max'' dictated the gating
decision) to be sometimes selected, thus allowing to accumulate a training
signal about the value of this configuration, i.e., a training signal
for the gater. The general question of {\em estimating or propagating gradients
through stochastic neurons} is treated in another exploratory
article~\citep{Bengio-arxiv2013}, where it is shown that one can obtain
an unbiased (but noisy) estimator of the gradient of a loss through a discrete
stochastic decision. Another interesting idea explored in that paper is
that of adding noise just before the non-linearity (max-pooling ($\max_i x_i$)
or rectifier ($\max(0,x)$)). Hence the winner
is not always the same, and when a choice wins it has a smooth influence
on the result, and that allows a gradient signal to be provided, pushing that
winner closer or farther from winning the competition on another example.

\section{Optimization}
\label{sec:optimization}

\subsection{Optimization: The Challenge}

As we consider larger and larger datasets (growing faster than the size of
the models), training error and generalization error converge.  Furthermore
many pieces of evidence in the results of experiments on deep learning
suggest that training deep networks (including recurrent networks) involves
a difficult
optimization~\citep{Bengio-chapter-2013,Gulcehre+Bengio-arxiv-2013,Bengio-et-al-ICASSP2013}. It
is not yet clear how much of the difficulty is due to local minima and how
much is due to ill-conditioning (the two main types of optimization
difficulties in continuous optimization problems).  It is therefore
interesting to study the optimization methods and difficulties involved in
deep learning, for the sake of obtaining better generalization.
Furthermore, better optimization could also have an impact on scaling
computations, discussed above.

One important thing to keep in mind, though, is that in a deep supervised
network, the top two layers (the output layer and the top hidden layer) can
rather easily be made to overfit, simply by making the top hidden layer
large enough.  However, to get good generalization, what we have found is
that one needs to {\em optimize the lower layers}, those that are far
removed from the immediate supervised training
signal~\citep{Bengio-nips-2006-small}.
These observations mean that only looking
at the training criterion is not sufficient to assess that a training
procedure is doing a good job at optimizing the lower layers well.
However, under constraints on the top hidden layer size, training error
can be a good guide to the quality of the optimization of lower layers.
Note that supervised deep nets are
very similar (in terms of the optimization problem involved) to deep
auto-encoders and to recurrent or recursive networks, and that properly
optimizing RBMs (and more so deep Boltzmann machines) seems more difficult: progress on
training deep nets is therefore likely to be a key to training the other
types of deep learning models.

One of the early hypotheses drawn from experiments with layer-wise
pre-training as well as of other experiments (semi-supervised
embeddings~\citep{WestonJ2008-small} and slow feature
analysis~\citep{wiskott:2002,Bergstra+Bengio-2009-small}) is that the
training signal provided by backpropagated gradients is sometimes too weak to
properly train intermediate layers of a deep network. This is supported by
the observation that all of these successful techniques somehow inject a
training signal into the intermediate layers, helping them to figure out
what they should do.  However, the more recent successful results with
supervised learning on very large labeled datasets suggest that with some
tweaks in the optimization procedure (including initialization), it is
sometimes possible to achieve as good results with or without unsupervised
pre-training or semi-supervised embedding intermediate training signals.

\subsection{Optimization: Solution Paths}

In spite of these recent encouraging results, several more recent
experimental results again point to a fundamental difficulty in training
intermediate and lower layers.

\subsubsection{Diminishing Returns with Larger Networks.}
First, \citet{Dauphin+Bengio-arxiv-2013} show that with
well-optimized SGD training, as the size of a neural net increases, the
``return on investment'' (number of training errors removed per added
hidden unit) decreases, given a fixed number of training iterations, until
the point where it goes below 1 (which is the return on investment that
would be obtained by a brain-dead memory-based learning mechanism -- such
as Parzen Windows -- which just copies an incorrectly labeled example into
the weights of the added hidden unit so as to produce just the right answer
for that example only). This suggests that larger models may be
fundamentally more difficult to train, probably because there are now more
second-order interactions between the parameters, increasing the condition
number of the Hessian matrix (of second derivatives of model parameters
with respect to the training criterion). This notion of return on investment
may provide a useful metric by which to measure the effect of different
methods to improve the scaling behavior of training and optimization
procedures for deep learning.

\subsubsection{Intermediate Concepts Guidance and Curriculum.}
Second, \citet{Gulcehre+Bengio-arxiv-2013} show that there are apparently
simple tasks on which standard black-box machine learning algorithms
completely fail. Even {\em supervised and pre-trained deep networks} were
tested and failed at these tasks.  These tasks have in common the
characteristic that the correct labels are obtained by the composition of
at least two levels of non-linearity and abstraction: e.g., the first level
involves the detection of objects in a scene and the second level involves
a non-linear logical operation on top of these (such as the detecting
presence of multiple objects of the same category).  On the other hand, the
task becomes easily solvable by a deep network whose intermediate layer is
first pre-trained to solve the first-level sub-task. This raises the
question of how humans might learn even more abstract tasks,
and \citet{Bengio-chapter-2013} studies the hypothesis that the use of
language and the evolution of culture could have helped humans reduce that
difficulty (and gain a serious advantage over other less cultured animals).
It would be interesting to explore multi-agent learning mechanisms inspired
by the the mathematical principles behind the evolution of culture in order
to bypass this optimization difficulty. The basic idea is that humans (and
current learning algorithms) are limited to ``local descent'' optimization
methods, that make small changes in the parameter values with the effect of
reducing the expected loss in average. This is clearly prone to the
presence of local minima, while a more global search (in the spirit of both
genetic and cultural evolution) could potentially reduce this
difficulty. One hypothesis is that more abstract learning tasks involve
more challenging optimization difficulties, which would make such global
optimization algorithms necessary if we want computers to learn such
abstractions from scratch.  Another option, following the idea of
curriculum learning~\citep{Bengio+al-2009-small}, is to provide guidance
ourselves to learning machines (as exemplified in the toy example
of \citet{Gulcehre+Bengio-arxiv-2013}), by ``teaching them'' gradually more
complex concepts to help them understand the world around us (keeping in
mind that we also have to do that for humans and that it takes 20 years to
complete).

\subsubsection{Changing the learning procedure and the architecture.}

Regarding the basic optimization difficulty of a single deep network, three types
of solutions should be considered. First, there are solutions based on
improved general-purpose optimization algorithms, such as for example the
recent work on adaptive learning rates~\citep{Schaul2012}, online
natural gradient~\citep{LeRoux+al-tonga-2008-small,Pascanu+Bengio-arxiv2013}
or large-minibatch second order methods~\citep{martens2010hessian}.

Another class of attacks on the optimization problem is based on changing
the architecture (family of functions and its parametrization) or the way
that the outputs are produced (for example by adding noise). As already
introduced in~\citet{LeCun+98backprop-small}, changes in the preprocessing,
training objective and architecture can change the difficulty of
optimization, and in particularly improve the conditioning of the Hessian
matrix (of second derivatives of the loss with respect to parameters).
With gradient descent, training time into a quadratic bowl is roughly
proportional to the condition number of the Hessian matrix (ratio of
largest to smallest eigenvalue).  For
example~\citet{LeCun+98backprop-small} recommends centering and normalizing
the inputs, an idea recently extended to hidden layers of Boltzmann
machines with success~\citep{Montavon2012}. A related idea that may have an
impact on ill-conditioning is the idea of skip-connections, which forces
both the mean output and the mean slope of each hidden unit of a deep
multilayer network to be zero~\citep{Raiko-2012-small}, a centering idea
which originates from~\citet{Schraudolph-1998}.

There has also been very successful recent work exploiting rectifier
non-linearities for deep supervised
networks~\citep{Glorot+al-AI-2011-small,Krizhevsky-2012-small}. Interestingly,
such non-linearities can produce rather sparse unit outputs, which could be
exploited, if the amount of sparsity is sufficiently large, to considerably
reduce the necessary computation (because when a unit output is 0, there is
no need to actually multiply it with its outgoing weights). Very
recently, we have discovered a variant on the rectifier non-linearity
called maxout~\citep{Goodfellow+al-ICML2013-small} which appears to open a
very promising door towards more efficient training of deep networks.  As
confirmed experimentally~\citep{Goodfellow+al-ICML2013-small}, maxout
networks can train deeper networks and allow lower layers to undergo more
training. The more general principle at stake here may be that when the
{\em gradient is sparse}, i.e., only a small subset of the hidden units and
parameters is touched by the gradient, the optimization problem may become
easier. We hypothesize that sparse gradient vectors have a positive effect
on reducing the ill-conditioning difficulty involved in training deep
nets. The intuition is that by making many terms of the gradient vector 0,
one also knocks off many off-diagonal terms of the Hessian matrix,
making this matrix more diagonal-looking, which would
reduce many of the ill-conditioning effects involved, as explained below.
Indeed, gradient descent relies on an invalid assumption: that one can
modify a parameter $\theta_i$ (in the direction of the gradient
$\frac{\partial C}{\partial \theta_i}$) without taking into account the
changes in $\frac{\partial C}{\partial \theta_i}$ that will take place when
also modifying other parameters $\theta_j$. Indeed, this is precisely the
information that is captured (e.g. with second-order methods) by the
off-diagonal entries $\frac{\partial^2
C}{\partial \theta_i \partial \theta_j} = \frac{\partial
}{\partial \theta_j} \frac{\partial C}{\partial \theta_i}$, i.e., how
changing $\theta_j$ changes the gradient on $\theta_i$. Whereas
second-order methods may have their own limitations\footnote{first,
practical implementations never come close to actually inverting the Hessian,
and second, they often require line searches that may be computationally inefficient
if the optimal trajectory is highly curved}
it would be
interesting if substantially reduced ill-conditioning could be achieved by
modifying the architecture and training procedure. Sparse gradients would
be just one weapon in this line of attack.

As we have argued above, adding noise in an appropriate way can be useful
as a powerful regularizer (as in dropouts), and it can also be used to make
the gradient vector sparser, which would reinforce the above positive
effect on the optimization difficulty.  If some of the activations are also
sparse (as our suggestions for conditional computation would require), then
more entries of the gradient vector will be zeroed out, also reinforcing
that beneficial optimization effect. In addition, it is plausible that the
masking noise found in dropouts (as well as in denoising auto-encoders)
encourages a faster {\em symmetry-breaking}: quickly moving away from
the condition where all hidden
units of a neural network or a Boltzmann machine do the same thing (due to
a form of symmetry in the signals they receive), which is a non-attractive fixed
point with a flat (up to several orders) likelihood function. This means
that gradient descent can take a lot of time to pull apart hidden units
which are behaving in a very similar way. Furthermore, when starting from
small weights, these symmetry conditions (where many hidden units do
something similar) are actually attractive from far away, because initially
all the hidden units are trying to grab the easiest and most salient job
(explain the gradients on the units at the layer above). By randomly
turning off hidden units we obtain a {\em faster specialization} which
helps training convergence.

A related concept that has been found useful in understanding and reducing
the training difficulty of deep or recurrent nets is the importance of {\em
letting the training signals (back-propagated gradients) flow, in a focused
way}.  It is important that error signals flow so that {\em credit and
blame is clearly assigned} to different components of the model, those
that could change slightly to improve the training loss. The problem of vanishing and
exploding gradients in recurrent nets~\citep{Hochreiter91,Bengio-trnn93}
arises because the effect of a long series of non-linear composition tends
to produce gradients that can either be very small (and the error signal is
lost) or very large (and the gradient steps diverge temporarily). This idea
has been exploited to propose successful initialization procedures for deep
nets~\citep{GlorotAISTATS2010-small}. A composition of non-linearities is
associated with a product of Jacobian matrices, and a way to reduce the
vanishing problem would be to make sure that they have a spectral radius
(largest eigenvalue) close to 1, like what is done in the weight
initialization for Echo State Networks~\citep{Jaeger-2007} or in the
carousel self-loop of LSTM~\citep{Hochreiter+Schmidhuber-1997} to help
propagation of influences over longer paths. A more generic way to avoid
gradient vanishing is to incorporate a training penalty that encourages the
propagated gradient vectors to maintain their
magnitude~\citep{Pascanu+Bengio-arxiv-2012}.  When combined with a gradient
clipping\footnote{When the norm of the gradient is above a threshold
$\tau$, reduce it to $\tau$} heuristic~\citep{Mikolov-thesis-2012} to avoid
the detrimental effect of overly large gradients, it allows to train
recurrent nets on tasks on which it was not possible to train them
before~\citep{Pascanu+Bengio-arxiv-2012}.

\section{Inference and Sampling}
\label{sec:inference}

All of the graphical models studied for deep learning except the humble RBM
require a non-trivial form of inference, i.e., guessing values of the
latent variables $h$ that are appropriate for the given visible input
$x$. Several forms of inference have been investigated in the past: MAP
inference is formulated like an optimization problem (looking for $h$ that
approximately maximizes $P(h\mid x)$); MCMC inference attempts to sample a
sequence of $h$'s from $P(h\mid x)$; variational inference looks for a
simple (typically factorial) approximate posterior $q_x(h)$ that is close
to $P(h\mid x)$, and usually involves an iterative optimization procedure.
See a recent machine learning textbook for more
details~\citep{bishop-book2006,Barber-2011,MurphyBook2012}.

In addition, a challenge related to inference is sampling (not just from
$P(h\mid x)$ but also from $P(h,x)$ or $P(x)$), which like inference is
often needed in the inner loop of learning algorithms for probabilistic
models with latent variables, energy-based models~\citep{lecun2006} or
Markov Random Fields~\citep{kindermann-book-1980} (also known as undirected
graphical models), where $P(x)$ or $P(h,x)$ is defined in terms of a
parametrized energy function whose normalized exponential gives
probabilities.

Deep Boltzmann machines~\citep{SalHinton09small} combine the challenge of
inference (for the {\em ``positive phase''} where one tries to push the
energies associated with the observed $x$ down) and the challenge of
sampling (for the {\em ``negative phase''} where one tries to push up the
energies associated with $x$'s sampled from $P(x)$).  Sampling for the
negative phase is usually done by MCMC, although some learning
algorithms~\citep{CollobertR2008-small,Gutmann+Hyvarinen-2010-small,Bordes-et-al-LSML2013}
involve ``negative examples'' that are sampled through simpler procedures
(like perturbations of the observed input). In \citet{SalHinton09small},
inference for the positive phase is achieved with a mean-field variational
approximation.\footnote{In the mean-field approximation, computation proceeds
like in Gibbs sampling, but with stochastic binary values replaced by their
conditional expected value (probability of being 1), given the outputs of
the other units. This deterministic computation is
iterated like in a recurrent network until convergence is approached, to
obtain a marginal (factorized probability) approximation over all the
units.}

\subsection{Inference and Sampling: The Challenge}

There are several challenges involved with all of the these inference
and sampling techniques.

The first challenge is practical and computational: these are all iterative
procedures that can considerably slow down training (because inference
and/or sampling is often in the inner loop of learning).

\subsubsection{Potentially Huge Number of Modes.}
The second challenge is more fundamental and has to do with the potential
existence of highly multi-modal posteriors: all of the currently known
approaches to inference and sampling are making very strong explicit or
implicit assumptions on the form the distribution of interest ($P(h\mid x)$
or $P(h,x)$). As we argue below, these approaches make sense if this target
distribution is either approximately unimodal (MAP), (conditionally)
factorizes (variational approximations, i.e., the different factors $h_i$
are approximately independent\footnote{this can be relaxed by considering
tree-structured conditional dependencies~\citep{Saul96}
and mixtures thereof} of each other given $x$), or has only a few
modes between which it is easy to mix (MCMC). However, approximate inference
can be potentially hurtful, not just at test time but for
training, because it is often in the inner loop of the learning 
procedure~\citep{Kulesza+Pereira-NIPS2007}.

Imagine for example that $h$
represents many explanatory variables of a rich audio-visual scene with a
highly ambiguous raw input $x$, including the presence of several objects
with ambiguous attributes or categories, such that one cannot really
disambiguate one of the objects independently of the others (the so-called
``structured output'' scenario, but at the level of latent explanatory
variables). Clearly, a factorized or unimodal representation would be
inadequate (because these variables are not at all independent, given $x$)
while the number of modes could grow
exponentially with the number of ambiguous factors present in the
scene. For example, consider a visual scene $x$ through a haze
hiding most details, yielding a lot of uncertainty. Say it involves 10 objects
(e.g., people), each
having 5 ambiguous binary attributes (out of 20)
(e.g., how they are dressed)
and uncertainty between 100
categorical choices for each element
(e.g., out of 10000 persons in the database, the marginal evidence allows
to reduce the uncertainty for each person to about 100 choices).
Furthermore, suppose that these uncertainties cannot be factorized
(e.g., people tend to be in the same room with other people involved
in the same activity, and friends tend to stand physically close to each other,
and people choose to dress in a way that socially coherent).
To make life hard on mean-field and other factorized approximations,
this means that only a small fraction (say 1\%) of these configurations
are really compatible.
So one really
has to consider $1\% \times (2^{5}\times
100)^{10} \approx 10^{33}$ {\em plausible configurations} of the latent
variables.
If one has to take a decision $y$ based on $x$, e.g.,
$P(y\mid x)=\sum_h P(y\mid h) P(h\mid x)$ involves summing over a huge
number of non-negligible terms of the posterior $P(h\mid x)$, which
we can consider as modes (the actual dimension of $h$ is much larger,
so we have reduced the problem from $(2^{20} \times 10000)^{10}\approx 10^{100}$
to about $10^{33}$, but that is still huge.
One way or another, {\em summing explicitly over that many
modes seems implausible}, and assuming single mode (MAP) or a factorized
distribution (mean-field) would yield very poor results.
Under some assumptions on the underlying data-generating process,
it might well be possible to do inference that is exact or a
provably good approximations, and searching for graphical models
with these properties is an interesting avenue to deal with this problem.
Basically, these assumptions work because we assume a specific structure
in the form of the underlying distribution.
Also, if we are lucky, a few Monte-Carlo
samples from $P(h\mid x)$ might suffice to obtain an acceptable approximation
for our $y$, because somehow, as far as $y$ is concerned, many probable values of $h$
yield the same answer $y$ and a Monte-Carlo sample will well represent these
different ``types'' of values of $h$.
That is one form of regularity that could be exploited (if it exists) to
approximately solve that problem. What if these assumptions are not
appropriate to solve challenging AI problems? Another, more general assumption (and thus
one more likely to be appropriate for these problems) is similar to what
we usually do with machine learning: although the space of functions is
combinatorially large, we are able to generalize by postulating a rather
large and flexible family of functions (such as a deep neural net).
Thus an interesting avenue is to assume that there exists a
computationally tractable function that can compute $P(y\mid x)$
in spite of the apparent complexity of going through the intermediate
steps involving $h$, and that we may learn $P(y\mid x)$ through $(x,y)$ examples.
This idea will be developed further in Section~\ref{sec:inference-solutions}.

\subsubsection{Mixing Between Modes.}
What about MCMC methods? They are hurt by the problem of mode mixing,
discussed at greater length in~\citet{Bengio-et-al-ICML2013}, and
summarized here.  To make the mental picture simpler, imagine that there
are only two kinds of probabilities: tiny and high. MCMC transitions try to
stay in configurations that have a high probability (because they should
occur in the chain much more often than the tiny probability
configurations). Modes can be thought of as islands of high probability,
but they may be separated by vast seas of tiny probability configurations.
Hence, it is difficult for the Markov chain of MCMC methods to jump from
one mode of the distribution to another, when these are separated by large
low-density regions embedded in a high-dimensional space, a common
situation in real-world data, and under the {\em manifold
hypothesis} \citep{Cayton-2005,Narayanan+Mitter-NIPS2010-short}.  This
hypothesis states that natural classes present in the data (e.g., visual
object categories) are associated with low-dimensional regions\footnote{e.g. they
can be charted with a few coordinates} (i.e., manifolds) near
which the distribution concentrates, and that different class manifolds are
well-separated by regions of very low density. Here, what we consider a
mode may be more than a single point, it could be a whole (low-dimensional)
manifold. Slow mixing between modes means that consecutive samples tend to
be correlated (belong to the same mode) and that it takes a very large
number of consecutive sampling steps to go from one mode to another and
even more to cover all of them, i.e., to obtain a large enough
representative set of samples (e.g. to compute an expected value under the
sampled variables distribution).  This happens because these jumps through
the low-density void between modes are unlikely and rare events. When a
learner has a poor model of the data, e.g., in the initial stages of
learning, the model tends to correspond to a smoother and higher-entropy
(closer to uniform) distribution, putting mass in larger volumes of input
space, and in particular, between the modes (or manifolds). This can be
visualized in generated samples of images, that look more blurred and
noisy\footnote{See examples of generated images with some of the current
state-of-the-art in learned generative models of
images \citep{Courville+al-2011-small,Luo+al-AISTATS2013-small}}.  Since
MCMCs tend to make moves to nearby probable configurations, mixing between
modes is therefore initially easy for such poor models.  However, as the
model improves and its corresponding distribution sharpens near where the
data concentrate, mixing between modes becomes considerably slower.  Making
one unlikely move (i.e., to a low-probability configuration) may be
possible, but making $N$ such moves becomes exponentially unlikely in $N$.
Making moves that are far {\em and probable} is fundamentally difficult in
a high-dimensional space associated with a peaky distribution (because the
exponentially large fraction of the far moves would be to an unlikely
configuration), unless using additional (possibly learned) knowledge about
the structure of the distribution.

\subsection{Inference and Sampling: Solution Paths}
\label{sec:inference-solutions}

\subsubsection{Going into a space where mixing is easier.}
The idea of {\em tempering}~\citep{Iba-2001} for MCMCs is analogous to the
idea of simulated annealing~\citep{Kirkpatrick83} for optimization, and it
is designed for and looks very appealing to solve the mode mixing problem:
consider a smooth version (higher temperature, obtained by just dividing
the energy by a temperature greater than 1) of the distribution of
interest; it therefore spreads probability mass more uniformly so one can
mix between modes at that high temperature version of the model, and then
gradually cool to the target distribution while continuing to make MCMC
moves, to make sure we end up in one of the ``islands'' of high
probability. \citet{Desjardins+al-2010-small,Cho10IJCNN-small,Salakhutdinov-2010-small,Salakhutdinov-ICML2010-small}
have all considered various forms of tempering to address the failure of
Gibbs chain mixing in RBMs. Unfortunately, convincing solutions (in the sense of
making a practical impact on training efficiency) have not yet been clearly
demonstrated. It is not clear why this is so, but it may be due to the need
to spend much time at some specific (critical) temperatures in order to
succeed. More work is certainly warranted in that direction.

An interesting observation~\citep{Bengio-et-al-ICML2013} which could turn
out to be helpful is that after we train a deep model such as a DBN or a
stack of regularized auto-encoders, we can observe that mixing between
modes is much easier at higher levels of the hierarchy (e.g. in the
top-level RBM or top-level auto-encoder): mixing between modes is easier at
deeper levels of representation.  This is achieved by running the MCMC in a
high-level representation space and then projecting back in raw input space
to obtain samples at that level. The hypothesis
proposed~\citep{Bengio-et-al-ICML2013} to explain this observation is that
unsupervised representation learning procedures (such as for the RBM and
contractive or denoising auto-encoders) tend to discover a representation
whose distribution has more entropy (the distribution of vectors in higher
layers is more uniform) and that better ``disentangles'' or separates out
the underlying factors of variation (see next section for a longer
discussion of the concept of disentangling). For example, suppose that a
perfect disentangling had been achieved that extracted the factors out of
images of objects, such as object category, position, foreground color,
etc.  A single Gibbs step could thus switch a single top-level
variable (like object category) when that variable is resampled given the
others, a very local move in that top-level disentangled representation but
a very far move (going to a very different place) in pixel space.  Note
that maximizing mutual information between inputs and their learned {\em
deterministic} representation, which is what auto-encoders basically
do~\citep{VincentPLarochelleH2008-small}, is equivalent to maximizing the
entropy of the learned representation,\footnote{Salah Rifai, personal
communication} which supports this hypothesis.  An interesting
idea\footnote{Guillaume Desjardins, personal communication} would therefore
be to use higher levels of a deep model to help the lower layers mix
better, by using them in a way analogous to parallel tempering, i.e., to
suggest configurations sampled from a different mode.

Another interesting potential avenue for solving the problem of 
sampling from a complex and rough (non-smooth) distribution would
be to take advantage of quantum annealing effects~\citep{Rose07}
and analog computing hardware (such as produced by D-Wave).
NP-hard problems (such as sampling or optimizing exactly in an Ising model)
still require exponential time but experimental evidence has shown that
for some problems, quantum annealing is far superior to standard
digital computation~\citep{Brooke2001}. Since quantum annealing
is performed by essentially implementing a Boltzmann machine in analog
hardware, it might be the case that drawing samples from a Boltzmann
machine is one problem where quantum annealing would be dramatically
superior to classical digital computing.

\subsubsection{Learning a Computational Graph that Does What we Want}

If we stick to the idea of obtaining actual values of the latent variables
(either through MAP, factorized variational inference or MCMC), then a
promising path is based on {\em learning approximate inference}, i.e.,
optimizing a learned approximate inference mechanism so that it performs a
better inference faster.  This idea is not new and has been shown to work
well in many settings.  This idea was actually already present in the
wake-sleep algorithm~\citep{Hinton95,Frey96,Hinton06} in the context of
variational inference for Sigmoidal Belief Networks and DBNs.
Learned approximate inference is also crucial in the
predictive sparse coding (PSD) algorithm~\citep{koray-psd-08-small}.  This
approach is
pushed further with~\citet{Gregor+LeCun-ICML2010-small} in which the
parametric encoder has the same structural form as a fast iterative sparse
coding approximate inference algorithm. The important consideration in both
cases is not just that we have {\em fast} approximate inference, but that (a) it is
learned, and (b) the model is learned jointly with the learned approximate
inference procedure.  See
also~\citet{Salakhutdinov+Larochelle-2010-small} for learned fast approximate
variational inference in DBMs, or
~\citet{Bradley+Bagnell-2009-small,Stoyanov2011-small} for learning
fast approximate inference (with fewer steps than would otherwise
be required by standard general purpose inference)
based on loopy belief propagation.

The traditional view of probabilistic graphical models is based on the
clean separation between modeling (defining the model), optimization
(tuning the parameters), inference (over the latent variables) and sampling
(over all the variables, and possibly over the parameters as well in the
Bayesian scenario). This modularization has clear advantages but may be
suboptimal. By bringing learning into inference and jointly learning the
approximate inference and the ``generative model'' itself, one can hope to
obtain ``specialized'' inference mechanisms that could be much more
efficient and accurate than generic purpose ones; this was the subject of a
recent ICML workshop~\citep{eisner-2012-icmlw}.  The idea of learned
approximate inference may help deal with the first (purely computational)
challenge raised above regarding inference, i.e., it may help to speed up
inference to some extent, but it generally keeps the approximate inference
parameters separate from the model parameters.

But what about the challenge from a huge number of modes?  What if the
number of modes is too large and/or these are too well-separated for MCMC
to visit efficiently or for variational/MAP inference to approximate
satisfactorily?  If we stick to the objective of computing actual values of
the latent variables, the logical conclusion is that we should learn to
approximate a posterior that is represented by a rich multi-modal
distribution. To make things concrete, imagine that we learn (or identify)
a function $f(x)$ of the visible variable $x$ that computes the parameters
$\theta=f(x)$ of an approximate posterior distribution $Q_{\theta=f(x)}(h)$
but where $Q_{\theta=f(x)}(h)\approx P(h\mid x)$ can be highly multi-modal,
e.g., an RBM with visible variables $h$ (coupled with {\em additional
latent variables} used only to represent the richness of the posterior over
$h$ itself). Since the parameters of the RBM are obtained through a
parametric computation taking $x$ as input,\footnote{for many models, such
as deep Boltzmann machines, or bipartite discrete Markov random
fields~\citep{Martens+Sutskever-2010-small}, $f$ does not even need to be
learned, it can be derived analytically from the form of $P(h\mid x)$} this
is really a {\em conditional RBM}~\citep{Taylor+2007,TaylorHintonICML2009}.
Whereas variational inference is usually limited to a non-parametric
approximation of the posterior, $Q(h)$ (one that is analytically and
iteratively optimized for each given $x$) one could consider a parametric
approximate posterior that is learned (or derived analytically) while
allowing for a rich multi-modal representation (such as what an RBM can
capture, i.e., up to an exponential number of modes).

\subsubsection{Avoiding inference and explicit marginalization over latent variables altogether.}

We now propose to consider an even more {\em radical departure from traditional
thinking regarding probabilistic models with latent variables}. It is
motivated by the observation that even with the last proposal, something
like a conditional RBM to capture the posterior $P(h\mid x)$, when one has
to actually make a decision or a prediction, it is necessary for optimal
decision-making to {\em marginalize over the latent variables}.  For
example, if we want to predict $y$ given $x$, we want to compute something
like $\sum_h P(y\mid h) P(h\mid x)$.  If $P(h\mid x)$ is complex and highly
multi-modal (with a huge number of modes), then even if we can {\em
represent} the posterior, performing this sum exactly is out of the
question, and even an MCMC approximation may be either very poor
(we can only visit at most $N$ modes with $N$ MCMC steps, and that is very
optimistic because of the mode mixing issue) or very slow (requiring an
exponential number of terms being computed or a very very long MCMC chain).
It seems that we have not really addressed the original ``fundamental
challenge with highly multi-modal posteriors'' raised above.

To address this challenge, we propose to avoid explicit inference
altogether by {\em avoiding to sample, enumerate, or represent} actual
values of the latent variables $h$. In fact, our proposal is to completely
skip the latent variables themselves. Instead, if we first consider the example
of the previous paragraph, one can just directly learn to
predict $P(y\mid x)$. In general, what we seek is that
{\em the only approximation error we are left with is due to
to function approximation}. This might be important because the compounding
of approximate inference with function approximation could be very
hurtful~\citep{Kulesza+Pereira-NIPS2007}.

To get there, one may wish to mentally go through an
intermediate step. Imagine we had a good approximate posterior
$Q_{\theta=f(x)}(h)$ as proposed above, with parameters $\theta=f(x)$.
Then we could imagine learning an approximate
decision model that approximates and skips the intractable sum over $h$,
instead directly going from $\theta=f(x)$ to a prediction of $y$, i.e., we
would estimate $P(y\mid x)$ by $g(f(x))$. Now since we are already learning
$f(x)$, why learn $g(\theta)$ separately? We could simply directly learn to
estimate $\pi(x)=g(f(x))\approx P(y\mid x)$.

Now that may look trivial, because this is already what we do in discriminant
training of deep networks or recurrent networks, for example. And don't we
lose all the advantages of probabilistic models, such as, handling different
forms of uncertainty,
missing inputs, and being able to answer any ``question'' of the form ``predict
any variables given any subset of the others''? Yes, if we stick to
the traditional deep (or shallow) neural networks like those discussed
in Section~\ref{sec:dnn}.\footnote{although, using something like these
deep nets would be appealing because they are currently beating benchmarks in
speech recognition, language modeling and object recognition} But there
are other options.

{\em We propose to get the advantages of probabilistic models without the need
for explicitly going through many configurations of the latent variables.}
The general principle of what we propose to achieve this is to {\em construct
a family of computational graphs which perform the family of tasks we are
interested in.} A recent proposal~\citep{Goodfellow+al-ICLR2013} goes in
this direction. Like previous work on learned 
approximate inference~\citep{Stoyanov2011-small}, one can view the approach
as constructing a computational graph associated to approximate inference 
(e.g. a fixed number of iterations of mean-field updates) in a particular
setting (here, filling missing input with a variational approximation over
hidden and unclamped inputs). An interesting property is that depending
on which input variables are clamped and which are considered missing
(either during training or at test time), we get a different computational
graph, while all these computational graphs share the same parameters.
In~\citet{Goodfellow+al-ICLR2013}, training the shared parameters of these computational graph 
is achieved through a variational criterion that is similar to
a generalized pseudo-likelihood, i.e., 
approximately maximizing $\log P(x_v\mid x_c)$ for randomly chosen partitions $(v,c)$ of $s$.

This would be
similar to {\em dependency networks}~\citep{HeckermanD2000}, but
re-using the same parameters for every possible question-answer partition
and training the system to answer for any subset of variables rather
than singletons like in pseudo-likelihood. For the same reason, it raises
the question of whether the different estimated conditionals are coherent
with a global joint distribution. In the case where the computational
graph is obtained from the template of an inference mechanism for a joint
distribution (such as variational inference), then clearly, we keep the
property that these conditionals are coherent with a global joint distribution.
With the
mean-field variational inference, the computational graph looks like a
recurrent neural network converging to a fixed point, and where we stop the
iterations after a fixed number of steps or according to a convergence
criterion. Such a trained parametrized computational graph
is used in the iterative variational approach introduced
in~\citet{Goodfellow+al-ICLR2013} for training and missing value inference in
deep Boltzmann machines, with an inpainting-like criterion in which
arbitrary subsets of pixels are predicted given the others (a generalized
pseudo-likelihood criterion).  It has also been used in a recursion that
follows the template of loopy belief propagation to fill-in the missing
inputs and produce outputs~\citep{Stoyanov2011-small}.  Although in these cases there is
still a notion of latent variables (e.g. the latent variables of the deep Boltzmann machine) that
motivate the ``template'' used for the learned approximate inference, what
we propose here is to stop thinking about them as actual latent factors,
but rather just as a way to parametrize this template for a question
answering mechanism regarding missing inputs, i.e., the ``generic
conditional prediction mechanism'' implemented by the recurrent computational
graph that is trained to predict any subset of variables given any other subset.
Although~\citet{Goodfellow+al-ICLR2013} assume a factorial distribution
across the predicted variables, we propose to investigate non-factorial
posterior distributions over the observed variables, i.e., in the spirit of
the recent flurry of work on {\em structured output} machine
learning~\citep{TsochantaridisI2005}.
We can think of this parametrized computational graph as a family of
functions, each corresponding to answering a different question
(predict a specific set of variables given some others), but all sharing the same parameters.
We already have examples of such families in machine learning, e.g., with
recurrent neural networks or dynamic Bayes nets (where the functions in the
family are indexed by the length of the sequence). This is also analogous
to what happens with dropouts, where we have an exponential number of
neural networks corresponding to different sub-graphs from input to output
(indexed by which hidden units are turned on or off).  For the same reason
as in these examples, we obtain a form of generalization across
subsets. Following the idea of learned approximate inference, the
parameters of the question-answering inference mechanism would be taking advantage
of the specific underlying structure in the data generating distribution.
Instead of trying to do inference on the anonymous latent variables, it would be
trained to do good inference only over observed variables or over high-level
features learned by a deep architecture, obtained deterministically from
the observed input.

An even more radically different solution to the problem of avoiding
explicit latent variables was recently introduced
in~\citet{Bengio-et-al-arxiv-2013}
and~\citet{Bengio+Laufer-arxiv-2013}. These introduce training criteria
respectively for generalized forms of denoising auto-encoders and for
generative stochastic networks, with the property that maximum likelihood
training of the reconstruction probabilities yields consistent but implicit
estimation of the data generating distribution. These Generative Stochastic
Networks (GSNs) can be viewed as inspired by the Gibbs sampling procedure
in deep Boltzmann machines (or deep belief networks) in the sense that one
can construct computational graphs that perform similar computation, i.e.,
these are stochastic computational graphs (or equivalently, deterministic
computational graphs with noise sources injected in the graph). These
models are not explicitly trained to fill-in missing inputs but simply to
produce a Markov chain whose asymptotic distribution estimates the data
generating distribution.  However, one can show that this chain can be
manipulated in order to obtain samples of the estimated conditional
distribution $P(x_v\mid x_c)$, i.e., if one clamps some of the inputs,
one can sample from a chain that stochastically fills-in from the
missing inputs.

The approximate inference is not anymore an approximation of something
else, it is the definition of the model itself. This is actually good news
because we thus eliminate the issue that the approximate inference may be
poor. The only thing we need to worry about is whether the parameterized
computational graph is rich enough (or may
overfit) to capture the unknown data generating distribution, and whether
it makes it easy or difficult to optimize the parameters.

The idea that we should train with the approximate inference as part of the
computational graph for producing a decision (and a loss)
was first introduced by~\citet{Stoyanov2011-small}, and we simply
push it further here, by proposing to allow the computational graph to
depart in any way we care to explore from the template provided by existing
inference or sampling mechanisms, i.e., potentially losing the connection and the
reference to probabilistic latent variables.
Once we free ourselves from the constraint of interpreting this
parametrized question answering computational graph as corresponding to
approximate inference or approximate sampling involving latent variables, all kinds of
architectures and parametrizations are possible, where current approximate
inference mechanisms can serve as inspiration and starting points. Interestingly,
\citet{Bengio+Laufer-arxiv-2013} provides a proper training criterion for
training any such stochastic computational graph {\em simply using backprop
over the computational graph}, so as to maximize the probability of reconstructing
the observed data under a reconstruction probability distribution that depends
on the inner nodes of the computational graph. The noise injected in the computational
graph must be such that the learner cannot get rid of the noise and obtain
perfect reconstruction (a dirac at the correct observed input), just like
in denoising auto-encoders.
It is
quite possible that this new freedom could give rise to much better
models. 

To go farther than ~\citet{Bengio+Laufer-arxiv-2013,Goodfellow+al-ICLR2013,Stoyanov2011-small}
it would be good to go beyond the kind of factorized prediction common
in variational and loopy belief propagation inference. We would like the
reconstruction distribution to be able to capture multi-modal non-factorial
distributions. Although the result from ~\citet{Alain+Bengio-ICLR2013} suggests
that when the amount of injected noise is small, a unimodal distribution is
sufficient, it is convenient to accomodate large amounts of injected noise
to make training more efficient, as discussed by \citet{Bengio-et-al-arxiv-2013}.
One idea is to obtain such multi-modal reconstruction distributions is to
represent the estimated joint distribution of the predicted variables
(possibly given the clamped variables) by a powerful model such as an RBM or
a regularized auto-encoder, e.g., as has been done for structured
output predictions when there is complex probabilistic structure between
the output variables~\citep{Mnih-2011,Li-et-al-CVPR2013}.

Although conditional RBMs have been already explored, conditional
distributions provided by regularized auto-encoders remain to be studied.
Since a denoising auto-encoder can be shown to estimate the underlying
data generating distribution, making its parameters dependent on some
other variables yields an estimator of a conditional distribution, which
can also be trained by simple gradient-based methods (and backprop to 
obtain the gradients).

All these ideas lead to the question: what is the interpretation of hidden
layers, if not directly of the underlying generative latent factors?  The
answer may simply be that they provide a better representation of these factors, a subject
discussed in the next section. But what about the representation of uncertainty
about these factors? The author believes that humans and other animals carry
in their head an internal representation that {\em implicitly} captures both the most likely
interpretation of any of these factors (in case a hard decision about some
of them has to be taken) and uncertainty about their joint assignment.
This is of course a speculation.
Somehow, our brain would be {\em operating on implicit representations of the
joint distribution between these explanatory factors}, generally without having to
commit until a decision is required or somehow provoked by
our attention mechanisms (which seem related
to our tendancy to verbalize a discrete interpretation). A good example is foreign language
understanding for a person who does not master that foreign language. Until we
consciously think about it, we generally don't
commit to a particular meaning for ambiguous word (which would be required
by MAP inference), or even to the segmentation of
the speech in words, but we can take a hard or a stochastic
decision that depends on the interpretation of these words if we have to, without
having to go through this intermediate step of discrete interpretation, instead treating
the ambiguous information as soft cues that may inform our decision.
In that example, a factorized posterior is also inadequate
because some word interpretations are more compatible with each other.

To summarize, what we propose here, unlike in previous work on
approximate inference, is to drop the pretense that the learned approximate
inference mechanism actually approximates the latent variables
distribution, mode, or expected value. Instead, we only consider the
construction of a computational graph (deterministic or stochastic) which
produces answers to the questions we care about, and we make sure
that we can train a family of computational graphs (sharing parameters)
whose elements can answer any of these questions. By removing the
interpretation of approximately marginalizing over latent variables, we
free ourselves from a strong constraint and the possible hurtful
approxiations involved in approximate inference, especially when the
true posterior would have a huge number of significant modes.

This discussion is of course orthogonal to the use of Bayesian
averaging methods in order to produce better-generalizing predictions,
i.e., handling uncertainty due to a small number of training examples.
The proposed methods can be made Bayesian just like neural networks
have their Bayesian variants~\citep{Neal94}, by somehow maintaining
an implicit or explicit distribution over parameters. A promising
step in this direction was proposed by~\citet{Welling+Teh-ICML2011}, making such Bayesian
computation tractable by exploiting the randomness introduced with
stochastic gradient descent to also produce the Bayesian samples over
the uncertain parameter values.

\section{Disentangling}

\subsection{Disentangling: The Challenge}

What are ``underlying factors'' explaining the data? The answer is not
obvious. One answer could be that these are factors that can be {\em separately
controlled} (one could set up way to change one but not the others).
This can actually be observed by looking at sequential real-world data,
where only a small proportion of the factors typically change from $t$ to $t+1$.
Complex data arise from the rich interaction of many
sources. These factors interact in a complex web that can complicate
AI-related tasks such as object classification. If we could identity and
separate out these factors (i.e., disentangle them), we would have almost
solved the learning problem. For example, an image is composed of the
interaction between one or more light sources, the object shapes and the
material properties of the various surfaces present in the image.  It is
important to distinguish between the related but distinct goals of learning
invariant features and learning to disentangle explanatory factors. The
central difference is the preservation of information. Invariant features,
by definition, have reduced sensitivity in the directions of
invariance. This is the goal of building features that are insensitive to
variation in the data that are uninformative to the task at
hand. Unfortunately, it is often difficult to determine \emph{a priori}
which set of features and variations will ultimately be relevant to the
task at hand. Further, as is often the case in the context of deep learning
methods, the feature set being trained may be destined to be used in
multiple tasks that may have distinct subsets of relevant features.
Considerations such as these lead us to the conclusion that the most robust
approach to feature learning is {\em to disentangle as many factors as
possible, discarding as little information about the data as is practical}.

Deep learning algorithms that can do a much better job of disentangling the
underlying factors of variation would have tremendous impact. For example,
suppose that the underlying factors can be ``guessed'' (predicted) from a
simple (e.g. linear) transformation of the learned representation, ideally
a transformation that only depends on a few elements of the representation.  That
is what we mean by a representation that disentangles the underlying factors. It
would clearly make learning a new supervised task (which may be related to
one or a few of them) much easier, because the supervised learning could
quickly learn those linear factors, zooming in on the parts of the
representation that are relevant.

Of all the challenges discussed in this paper, this is probably
the most ambitious, and success in solving it the most likely
to have far-reaching impact. In addition to the obvious observation
that disentangling the underlying factors is almost like pre-solving
any possible task relevant to the observed data, having disentangled
representations would also solve other issues, such as the issue
of mixing between modes. We believe that it would also considerably
reduce the optimization problems involved when new information arrives
and has to be reconciled with the world model implicit in the
current parameter setting. Indeed, it would allow only changing
the parts of the model that involve the factors that are relevant
to the new observation, in the spirit of {\em sparse updates}
and {\em reduced ill-conditioning} discussed above.

\subsection{Disentangling: Solution Paths}

\subsubsection{Deeper Representations Disentangle Better.}

There are some encouraging signs that our current unsupervised
representation-learning algorithms are reducing the ``entanglement'' of the
underlying factors\footnote{as measured by how predictive some individual
features are of known factors} when we apply them to raw data (or to the
output of a previous representation learning procedure, like when we stack
RBMs or regularized auto-encoders).

First, there are experimental observations suggesting that sparse
convolutional RBMs and
sparse denoising auto-encoders achieve in their hidden units a greater
degree of disentangling than in their
inputs~\citep{Goodfellow2009,Glorot+al-ICML-2011-small}.  What these
authors found is that some hidden units were particularly sensitive to a
known factor of variation while being rather insensitive (i.e., invariant)
to others. For example, in a sentiment analysis model that sees unlabeled
paragraphs of customer comments from the Amazon web site, some hidden units
specialized on the topic of the paragraph (the type of product being
evaluated, e.g., book, video, music) while other units specialized on the
sentiment (positive vs negative). The disentanglement was never perfect, so
the authors made quantitative measurements of sensitivity and invariance
and compared these quantities on the input and the output (learned
representation) of the unsupervised learners.

Another encouraging observation (already mentioned in the section on
mixing) is that deeper representations were empirically found to be more
amenable to quickly mixing between modes~\citep{Bengio-et-al-ICML2013}. Two
(compatible) hypotheses were proposed to explain this observation: (1) RBMs
and regularized auto-encoders deterministically transform\footnote{when
considering the features learned, e.g., the $P(h_i=1\mid x)$, for RBMs}
their input distribution into one
that is more uniform-looking, that better fills the space (thus creating easier
paths between modes), and (2) these algorithms tend to discover
representations that are more disentangled. The advantage of a higher-level
disentangled representation is that a small MCMC step (e.g. Gibbs) in that
space (e.g. flipping one high-level variable) can move in one step from one
input-level mode to a distant one, e.g., going from one shape / object
to another one, adding or removing glasses on the face of a person
(which requires a very sharp coordination of pixels far from each other
because glasses occupy a very thin image area), or replacing foreground
and background colors (such as going into a ``reverse video'' mode).

Although these observations are encouraging, we do not yet have a clear
understanding as to {\em why} some representation algorithms tend to move
towards more disentangled representations, and there are other experimental
observations suggesting that this is {\em far from sufficient}. In
particular,
\citet{Gulcehre+Bengio-arxiv-2013} show an example of a task on which
deep supervised nets (and every other black-box machine learning algorithm tried) fail,
on which a completely disentangled input representation makes the task
feasible (with a maxout network~\citep{Goodfellow+al-ICML2013-small}).
Unfortunately, unsupervised pre-training
applied on the raw input images failed to produce enough disentangling to
solve the task, even with the appropriate convolutional structure. What is interesting
is that we now have a simple artificial task on which we can evaluate new
unsupervised representation learning methods for their disentangling ability.
It may be that a variant of the
current algorithms will eventually succeed at this task, or it may be that
altogether different unsupervised representation learning algorithms are needed.

\subsubsection{Generic Priors for Disentangling Factors of Variation.}

A general strategy was outlined in
~\citet{Bengio-Courville-Vincent-TPAMI2013} to enhance the discovery of
representations which disentangle the underlying and unknown factors of
variation: it relies on exploiting {\em priors} about these factors. We are
most interested in {\em broad generic priors} that can be useful for a
large class of learning problems of interest in AI. We list these priors
here:

\noindent
$\bullet$ {\bf Smoothness}: assumes the function $f$ to be learned is
s.t. $x\approx y$ generally implies $f(x)\approx f(y)$. This most basic
prior is present in most machine learning, but is insufficient to get
around the curse of dimensionality.\\
$\bullet$ {\bf Multiple explanatory factors}: the data generating
distribution is generated by different underlying factors, and for the most
part what one learns about one factor generalizes in many configurations of
the other factors. The objective is to recover or at least disentangle
these underlying factors of variation. This assumption is behind the idea
of {\bf distributed representations}. More specific priors on the form of
the model can be used to enhance disentangling, such as multiplicative
interactions between the
factors~\citep{tenenbaum00separating,Desjardins-2012} or orthogonality of
the features derivative with respect to the
input~\citep{Dauphin-et-al-NIPS2011-small,Rifai2012,Sohn-et-al-ICML2013}. The parametrization
and training procedure may also be used to disentangle discrete
factors (e.g., detecting a shape) from associated continuous-valued
factors (e.g., pose parameters), as in transforming auto-encoders~\citep{Hinton-transforming-aa-2011-small},
spike-and-slab RBMs with pooled slab variables~\citep{Courville+al-2011-small}
and other pooling-based models that learn a feature subspace~\citep{Kohonen1996,Hyvarinen2000}.\\
$\bullet$ {\bf A hierarchical organization of explanatory factors}: the
concepts that are useful for describing the world around us can be defined
in terms of other concepts, in a hierarchy, with more {\bf abstract}
concepts higher in the hierarchy, defined in terms of less abstract ones.
This assumption is exploited with deep representations. Although stacking
single-layer models has been rather successful, much remains to be done
regarding the joint training of all the layers of a deep unsupervised model.\\
$\bullet$ {\bf Semi-supervised learning}: with inputs $X$ and target $Y$ to
predict, given $X$, a subset of the factors explaining $X$'s distribution
explain much of $Y$, given $X$. Hence representations that are useful for
spelling out $P(X)$ tend to be useful when learning $P(Y\mid X)$, allowing
sharing of statistical strength between the unsupervised and supervised
learning tasks. However, many of the factors that explain $X$ may dominate
those that also explain $Y$, which can make it useful to incorporate observations
of $Y$ in training the learned representations, i.e., by semi-supervised
representation learning.\\
$\bullet$ {\bf Shared factors across tasks}: with many $Y$'s of
interest or many learning tasks in general, tasks (e.g., the corresponding $P(Y\mid X,{\rm task})$)
are explained by factors that are shared with other tasks, allowing sharing
of statistical strength across tasks, e.g. for
multi-task and transfer learning or domain adaptation. This can be achieved
by sharing embeddings or representation functions
across tasks~\citep{CollobertR2008-small,Bordes-et-al-LSML2013}.\\
$\bullet$ {\bf Manifolds}: probability mass concentrates near regions that
have a much smaller dimensionality than the original space where the data
lives.  This is exploited with regularized auto-encoder
algorithms, but training criteria that would explicitly take
into account that we are looking for a concentration of mass in an integral
number directions remain to be developed.\\
$\bullet$ {\bf Natural clustering}: different values of categorical
variables such as object classes are associated with separate
manifolds. More precisely, the local variations on the manifold tend to
preserve the value of a category, and a linear interpolation between
examples of different classes in general involves going through a low
density region, i.e., $P(X\mid Y=i)$ for different $i$ tend to be well
separated and not overlap much. For example, this is exploited in the
Manifold Tangent Classifier~\citep{Dauphin-et-al-NIPS2011-small}.  This
hypothesis is consistent with the idea that humans have {\em named}
categories and classes because of such statistical structure (discovered by
their brain and propagated by their culture), and machine learning tasks
often involves predicting such categorical variables.\\
$\bullet$ {\bf Temporal and spatial coherence}:
this prior introduced in~\cite{Becker92} is similar to the natural
clustering assumption but concerns
sequences of observations: consecutive (from a sequence) or spatially
nearby observations tend to be {\em easily predictable from each other}.
In the special case typically studied, e.g., slow feature analysis~\citep{WisSej2002},
one assumes that consecutive values are close to each other, or that
categorical concepts remain either present or absent for most of the transitions.
More generally, different underlying factors change
at different temporal and spatial scales, and this could be exploited
to sift different factors into different categories based on their
temporal scale.\\
$\bullet$ {\bf Sparsity}: for any given observation $x$, only a small
fraction of the possible factors are relevant. In terms of representation,
this could be represented by features that are often zero (as initially
proposed by~\citet{Olshausen+Field-1996}), or more generally
by the fact that most of the
extracted features are {\em insensitive} to small variations of $x$.  This
can be achieved with certain forms of priors on latent variables (peaked at
0), or by using a non-linearity whose value is often flat at 0 (i.e., 0 and
with a 0 derivative), or simply by penalizing the magnitude of the
derivatives of the function mapping input to representation. A variant on
that hypothesis is that for any given input, only a small part of the model
is relevant and only a small subset of the parameters need to be updated.\\
$\bullet$ {\bf Simplicity of Factor Dependencies}: in good high-level
representations, the factors are related to each other through simple,
typically linear, dependencies.  This can be seen in many laws of physics,
and is assumed when plugging a linear predictor on top of a learned
representation.

\section{Conclusion}

Deep learning and more generally representation learning are recent areas
of investigation in machine learning and recent years of research have
allowed to clearly identify several major challenges for approaching the
performance of these
algorithms from that of humans. We have broken down these
challenges into four major areas: scaling computations, reducing the
difficulties in optimizing parameters,
designing (or avoiding) expensive
inference and sampling, and helping to learn representations that
better disentangle the unknown underlying factors of variation.  There is room for exploring
many paths towards addressing all of these issues, and we have presented
here a few appealing directions of research towards these challenges.

\subsection*{Acknowledgments}

The author is extremely grateful for the feedback and discussions he
enjoyed with collaborators Ian Goodfellow, Guillaume Desjardins,
Aaron Courville, Pascal Vincent, Roland Memisevic and Nicolas Chapados, which
greatly contributed to help form the ideas presented here and fine-tune
this manuscript.  He is also grateful for the funding support from NSERC,
CIFAR, the Canada Research Chairs, and Compute Canada.

{\small
\bibliography{strings,strings-shorter,ml,aigaion-shorter}
\bibliographystyle{natbib}
}

\end{document}